\documentclass[a4paper]{article}

\usepackage{INTERSPEECH2019}
\usepackage{multirow}
\usepackage[utf8]{inputenc}
\usepackage[T1]{fontenc}
\usepackage{color}
\definecolor{darkspringgreen}{rgb}{0.09, 0.45, 0.27}
\definecolor{bronze}{rgb}{0.8, 0.5, 0.2}
\newcommand{\DZ}[1]{}
\newcommand{\KH}[1]{}
\newcommand{\KHR}[1]{}
\newcommand{\TB}[1]{}
\newcommand{\GP}[1]{}
\newcommand{\RP}[1]{}
\newcommand{\TS}[1]{}
\usepackage[final]{changes}

\title{Phoneme-Based Contextualization for Cross-Lingual \\ Speech Recognition in End-to-End Models}
\name{Ke Hu$^{\star}$\thanks{$^\star$These authors contributed equally to this work. We would like to thank David Rybach for helpful comments and suggestions.}, Antoine Bruguier$^{\star}$, Tara N. Sainath, Rohit Prabhavalkar, Golan Pundak}
\address{
  Google, LLC, USA}
\email{\{huk,tonybruguier,tsainath,prabhavalkar,golan\}@google.com}

\begin{document}

\maketitle

\begin{abstract}
\GP{Suggested first paragraph: Contextual automatic speech recognition (ASR), that biases recognition towards a given context (e.g. user’s playlists, or contacts) is challenging in end-to-end (E2E) models, since such models have been found to recognize rare proper names poorly.}\KH{Following sentence rephrased.}
Contextual automatic speech recognition, i.e., biasing recognition towards a given context (e.g. user’s playlists, or contacts), is challenging in end-to-end (E2E) models.
Such models maintain a limited number of candidates during beam-search decoding, and have been found to recognize rare named entities poorly. The problem is exacerbated when biasing towards proper nouns in foreign languages, e.g., geographic location names, which are virtually unseen in training and are thus out-of-vocabulary (OOV).
While  grapheme  or  wordpiece  E2E  models  might have a difficult time spelling OOV words, phonemes are more acoustically salient and past work has shown that E2E phoneme models can better predict such words.
In this work, we propose an E2E model containing both English wordpieces and phonemes in the modeling space, and perform contextual biasing of foreign words at the phoneme level by mapping pronunciations of foreign words into similar English phonemes.
In experimental evaluations, we find that the proposed approach performs 16\% better than a grapheme-only biasing model, and 8\% better than a wordpiece-only biasing model on a foreign place name recognition task, with only slight degradation on regular English tasks.

\end{abstract}

\section{Introduction}


End-to-end (E2E) models have attracted increasing attention recently. Instead of building an automatic speech recognition (ASR) system from different components such as the acoustic model (AM), language model (LM), and pronunciation model (PM), E2E models rely on a single neural network to directly learn speech-to-text mapping. Representative systems include a word-based connectionist temporal classification (CTC) model \cite{soltau16}, recurrent neural network transducer (RNN-T) \cite{graves12, rao17}, and attention-based models such as ``Listen, Attend, and Spell" (LAS) \cite{chan16}. Recent advances have shown that E2E models can outperform the state-of-the-art conventional system when trained on thousands of hours of data \cite{chiu18, he18}.

In previous work~\cite{aleksic15}, it has been shown that contextual information (i.e., phrases relevant to recognition in the current context such as contact names, geographic place names, songs, etc.) can improve ASR accuracy. Such phrases are often foreign words, or are rarely seen in training. \deleted{Context information usually consists of a list of relevant phrases for recognition, for example, contact names, geographic place names, and song names \cite{aleksic15}. They are often rare phrases or even foreign words\deleted{,} which \replaced{are}{might be} seen infrequently in training. }\added{Recognizing these phrases is challenging. Conventional ASR systems model them as independent contextual LM using an n-gram weighted finite state transducer (WFST), and compose it with a baseline LM for on-the-fly (OTF) rescoring \cite{aleksic15, hall15}}. This idea is extended to a LAS model in \cite{williams18}, where an n-gram LM and a word-to-grapheme ``speller" are composed to produce a contextual FST \deleted{which was then used }for \replaced{rescoring}{the log probabilities of}. The approach is similar to shallow fusion \cite{kannan18} which interpolates E2E model scores with an external LM in beam search. To bring more relevant words for biasing in E2E models, \cite{pundak18} proposes to push biasing weights to each subword unit and \added{deals with} over-biasing\deleted{by adding failure arcs}. \RP{I'm not sure it will be obvious what `failure arcs' are in this context. Since this is just in the introduction, maybe it's better to remove that bit?}\KHR{Removed.}Further improvements such as biasing before beam pruning, and wordpiece-based biasing\deleted{, and increasing proper noun coverage,} have been proposed to achieve state-of-the-art biasing results \cite{schuster2012japanese, he18, zhao19}. \deleted{This also leads to significant improvement in biasing on sets with high OOV rates. }\deleted{In addition, prefixes are used to guide biasing to prevent regression on no-biasing scenarios.}\RP{We should cite Schuster and Nakajima in the context of WPM here}\KHR{Makes sense.}Another class of contextual biasing uses an all-neural approach. Contextual-LAS (CLAS)\deleted{,} is proposed in \cite{pundak18} to use a bias encoder to model contextual phrases as embeddings and shows significant improvement than OTF rescoring \cite{hall15}. Phonetic information has been incorporated to CLAS to improve rare word recognition \cite{bruguier19}.

Although biasing is improved by these techniques, they do not address \deleted{the issue of }cross-lingual recognition\deleted{, which is critical for applications such as navigation queries with foreign place names}. \replaced{In \cite{patel18}, }{For conventional models,}contextual biasing has been used to assist recognition of foreign words\deleted{ in \cite{patel18}}. With the phoneme mapping from a foreign language phoneme set to the recognizer's phoneme set, foreign words are modeled as a phoneme-level contextual FST for biasing. \deleted{Phoneme outputs are transduced to foreign words using a lexicon in decoding.}It is unclear whether such an approach can be directly applied to E2E models. Phoneme-only E2E systems have been shown to have inferior performance compared to grapheme or wordpiece models (WPM) in general\deleted{ no-biasing scenarios} \cite{sainath17, irie19}, but \deleted{However, \cite{sainath17} shows a phoneme-alone LAS model }shows better recognition of rare words and proper nouns.

\GP{Here is what I feel is missing: This work has the following novel components: 1. WPM + phones as output units. 2. Sampling of WPM / phones in training. 3. FST  based Biasing that is allowed to consume phones.}\KH{The following paragraph is further modified.}\replaced{In this work we propose}{The focus of the present work is} to incorporate phonemes to a wordpiece E2E model as modeling units\deleted{ to build an E2E wordpiece-phoneme model} and use phoneme-level FST for contextual biasing. \replaced{We propose a word-frequency based sampling strategy to randomly tokenize rare words into phonemes in the target sequence using a lexicon.}{To better model rare words, we introduce their phonemes (from a lexicon) as the target sequence in addition to wordpieces.} \deleted{On the other hand, regular words are modeled as wordpieces alone. We use word frequencies to determine whether a word is rare (see details in Section \ref{sec:wpp}). }This approach also mitigates accuracy regressions that have been observed when using phoneme-only E2E models~\cite{sainath17,irie19}.\RP{I would suggest replacing the previous sentence with ``This approach also mitigates accuracy regressions that have been observed when using phoneme-only E2E models~\cite{sainath17,irie19}"}\KHR{Done.} We train our model using only American English data and thus its wordpiece\replaced{s}{ model} and phoneme set (no data from foreign languages). In inference, given a list of foreign words, we bias the recognition using an English phoneme-level biasing FST, which is built by first tokenizing the words into foreign phonemes\deleted{ sequence using a foreign lexicon} and then mapping them to English phonemes using \cite{patel18}. For example, given a navigation query ``directions to Créteil" and the assumption that the French word ``Créteil" is in our biasing list, ``Créteil" is first tokenized to French phonemes as ``\texttt{k R e t E j}"\deleted{ using a French lexicon}, and then mapped to English phonemes ``\texttt{k r\textbackslash~E t E j}" for biasing\footnote{X-SAMPA notations are used for phonemes.}. \RP{Consider adding a footnote that these are XSAMPA phonemes.}\KHR{Done.} The phoneme mapping is necessary since our modeling units contain only English phonemes. \deleted{An X-SAMPA phoneme set is used as in \cite{patel18}. }In decoding, we propose to incorporate the pronunciation FST of the biasing words to consume English phoneme symbols and produce foreign words, using the aforementioned foreign lexicon and phoneme mapping, i.e. ``\texttt{k r\textbackslash~E t E j}" $\rightarrow$ Créteil (details in Section \ref{sec:phone_mapping} and \ref{sec:decode}). Wordpiece outputs are concatenated to form words. In experimental evaluations, we find that the proposed phoneme-based biasing using wordpiece-phoneme model successfully recognizes foreign words. It performs 16\% relatively better in terms of WER than the grapheme-only biasing model, and 8\% better than the wordpiece-only biasing model in a task of recognizing navigation queries containing French place names. The proposed model also has the advantage that it can be directly applied to other foreign languages for biasing without model scalability issues.


\section{Prior Work}
\label{sec:prior}
\subsection{Shallow Fusion E2E Biasing}
\label{sec:shallow_fusion_e2e_biasing}
\deleted{Context information usually refers to a list of phrases in different scenarios, e.g. contacts, apps and locations,}
Shallow fusion has been used in E2E models for decoding \cite{kannan18} and contextual biasing \cite{he18}. Biasing phrases are first represented as n-gram WFST in the word level ($G$), and then left composed with a ``speller" FST ($S$) to produce a contextual LM: $C=min(det(S \circ G))$. The speller transduces a sequence of subword units\deleted{, which comprise the modeling space of the E2E model,} to corresponding words.
\deleted{Typical subword units are wordpieces and graphemes \cite{he18}.} The contextual LM ($C$) is then used to rescore the log-probability outputs of the E2E model during beam search:

\begin{equation}
\vec{y}^* = \underset{\vec{y}}{\arg\max} \log P(\vec{y}|\vec{x}) + \lambda \log P_{C}(\vec{y})
\label{eq:biasing}
\end{equation}
\noindent
Here, $\vec{x}$ denotes acoustic observations, and $\vec{y}$ the subword-unit sequence. \added{$P$ is the probability estimation from the E2E model, and $P_C$ is the biasing rescoring probability.} $\lambda$ controls the weight of contextual LM in rescoring.

For E2E biasing, \cite{pundak18} explored biasing at beginning of word, end of word, and at subword units. The authors find that unlike biasing at the end of word in conventional models, biasing at subword units with weight pushing prevents candidates from being pruned early from the beam. \deleted{To prevent a partial prefix match, failure arcs are added to cancel the weighting in over-biasing scenarios. }In \cite{zhao19}, wordpieces have been shown to outperform graphemes in biasing since they create a sparser match of biasing units. All these improvements lead to significantly better biasing which is comparable to the state-of-the-art conventional model \cite{he18}. \TB{Rohit says this sentence is unclear and should be fixed. I agree (because I don't understand it either).} \KH{Rephrased.}\replaced{To avoid over-biasing}{For no-biasing scenarios}, \cite{zhao19} also proposed to \replaced{only activate biasing phrases when they are proceeded by a set of prefixes}{use prefixes to guide biasing}\deleted{and shows only slight degradation in no-biasing scenarios}.

\subsection{Phoneme Mapping}
Cross-lingual phoneme mapping has been used in conventional systems for recognizing foreign words \cite{patel18}. First, a phoneme mapping is learned by aligning the pronunciations between foreign and target languages using TTS-synthesized audio and a pronunciation learning algorithm \cite{bruguier17}. In inference, foreign words are first built to an FST ($G$). \replaced{Lexica}{Lexicons} ($L$) are constructed between target-language phonemes and foreign words using phoneme mapping, and then left composed with $G$ to construct a dynamic class LM for decoding:
\begin{equation}
G_d^{'}=Det(L)\circ G_d
\label{eq:phone_mapping}
\end{equation}
\noindent
where $d$ denotes a dynamic class label. In Section \ref{sec:phone_mapping}, we describe how phoneme mapping is incorporated to \replaced{a}{an} wordpiece-phoneme E2E model for contextual biasing.

\section{Phoneme-Based Biasing}
\label{sec:proposed}

The focus of this work is to bias toward rare cross-lingual words which are typically missing from the training set. We propose to do that by utilizing phonemes, which are not affected by orthography.
\replaced{Specifically, we}{However, \cite{sainath17} shows that an all-phoneme model degrades performance on common words compared to a grapheme model \cite{sainath17}, and is thus not a desirable choice.
An alternative would be to} augment the wordpiece modeling space of an E2E model with phonemes to train a wordpiece-phoneme model.

\subsection{Wordpiece-Phoneme Model}
\label{sec:wpp}
A wordpiece-phoneme model differs from a wordpiece-only model in that it may decompose a few words to phonemes in training. The output of the model is a single softmax whose symbol set is the union of wordpiece and phoneme symbols.\deleted{Proper nouns and rare words are good choices since they have been shown to be better recognized by phoneme models in \cite{sainath17}. Note that we use context-independent phonemes as in \cite{sainath17}.} We use a pronunciation lexicon to obtain phoneme sequences of words.\deleted{The lexicon contains words and their frequencies from training data, and is trimmed by removing homophones (e.g. ``flower" and ``flour"), homographs (e.g. ``live" as a verb or adjective), and pronunciation variants (e.g. ``either"). It thus only contain entries that are unambiguous when going from spelling to pronunciation or the other way around.}\deleted{We do not generate phonemes for out-of-lexicon words using a trained grapheme-to-phoneme model. The intuition is that when pronunciation is not known, it is simpler and cleaner to let the E2E model infer the pronunciation rather than bring in another independently trained model. In addition, we use the written form of a transcript and do not use any verbalizer. Thus, words like ``\$9.95" were never presented as phonemes.} Since phonemes show strength in recognizing rare words \cite{sainath17}, we want to present these words as phonemes more often. In a target sentence, we decide to randomly present the $i^{th}$ word as phonemes with a probability $p(i) = p_0 . \min ( \frac{T}{c(i)}, 1.0)$ where $p_0$ and $T$ are constants and $c(i)$ is an integer representing the number of time the word appears in our entire training corpus. Therefore, the words that appear $T$ times or less will be presented as phonemes with probability $p_0$. For words that appear more than $T$ times, the more frequent they are, the less likely they are \deleted{to be }presented as phonemes\footnote{Empirically, we find $T=10$ and $p_0=0.5$ to be reasonable choices, and higher $T$ and $p_0$ do not improve biasing significantly and may cause significant regressions to no-bias scenarios.}. Note that the decision of whether to use wordpieces or phonemes is made randomly at each gradient iteration, and thus a given sentence could have different target sequences at different epochs. \added{We use context-independent phonemes as in \cite{sainath17}.}

\subsection {Biasing FST for Phonemes}
\label{sec:phone_mapping}

\begin{figure}[h]
  \centering
  \includegraphics[scale=0.45]{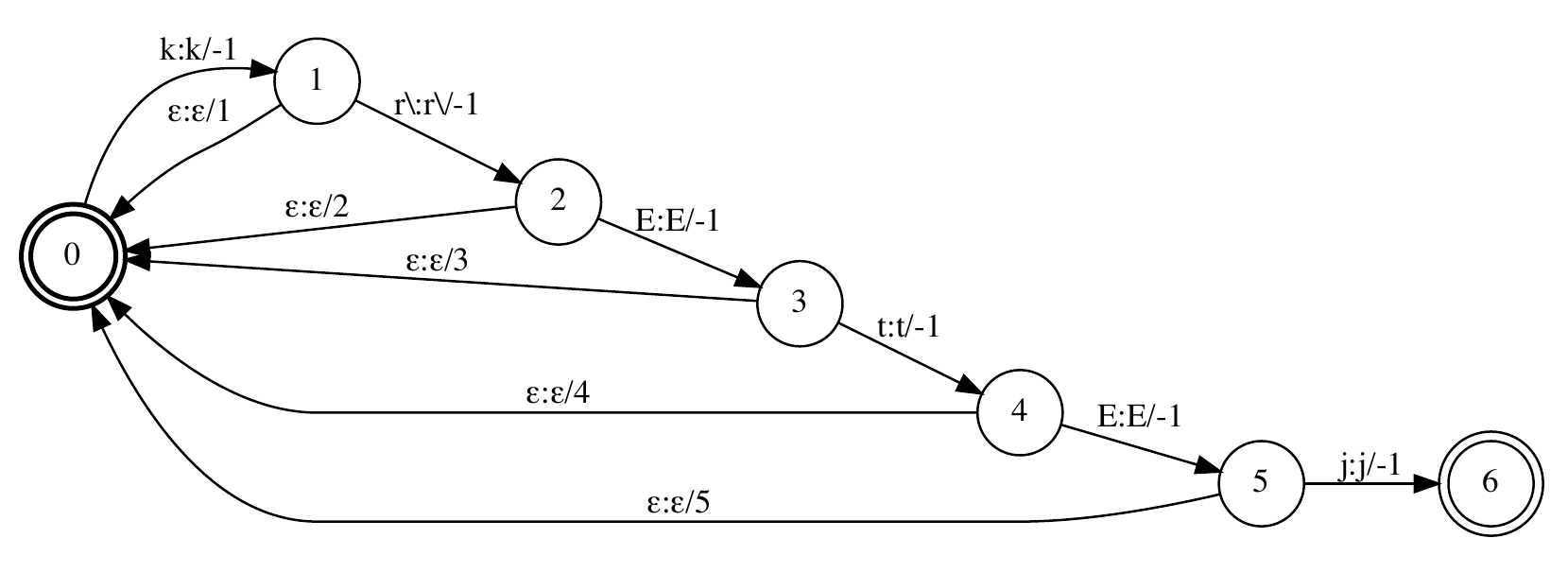}
   \caption{Contextual FST for the word ``Créteil" using a sequence of English phonemes ``\texttt{k r\textbackslash~E t E j}".}
   \label{fig:fst}
   \vspace{-1em}
\end{figure}

\added{In inference, cross-lingual biasing words are converted to English phonemes to rescore the phoneme outputs of the wordpiece-phoneme model.}\deleted{We investigate how to incorporate phoneme mapping to an English E2E wordpiece-phoneme model for phoneme-level biasing.} In our work, phoneme mapping is represented by a dictionary which contains \added{human generated} source-language to target-language phoneme pairs \cite{patel18}, and the X-SAMPA phoneme set is used for all languages.
For example, given a French word ``Créteil", we tokenize it into phonemes using the French pronunciation lexicon, i.e. ``Créteil" $\rightarrow$ ``\texttt{k R e t E j}", and then map the French phonemes to English phonemes one by one: ``\texttt{k R e t E j}" $\rightarrow$ ``\texttt{k r\textbackslash~E t E j}". Note that the mapping is needed since our wordpiece-phoneme model contains only English phonemes.\TB{Rohit says that the sentence needs to be explained since the reference is not published.}\KH{done} The English phoneme sequence is then used to construct a phoneme-level FST for biasing. Weight pushing is used to assign weights at the phoneme level and failure arcs are added to avoid over-biasing similar to \cite{pundak18}. Figure \ref{fig:fst} shows an example of a contextual FST for the word ``Créteil" at the phoneme level.\deleted{As described in Section \ref{sec:shallow_fusion_e2e_biasing}, Figure \ref{fig:fst} shows the biasing FST for the foreign word Créteil.} The biasing FST is then used to rescore the phoneme outputs of the wordpiece-phoneme model on the fly, using Eq. (\ref{eq:biasing}).

\subsection{Decoding Graph}
\label{sec:decode}

\begin{figure*}[t]
  \centering
  \includegraphics[scale=0.75]{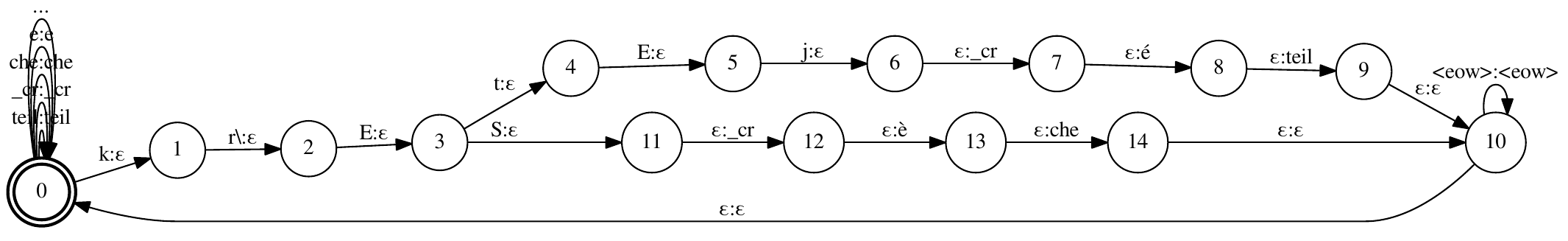}
  \caption{Decoding graph for the words ``crèche" (daycare) with English cross lingual pronunciation ``\texttt{k r\textbackslash ~E S}" and ``créteil" (a city) with pronunciation ``\texttt{k r\textbackslash ~E t E j}". For clarity, we omitted most wordpieces for the state 0.}
  \label{fig:decoding_fst}
  \vspace{-1em}
\end{figure*}

To generate words as outputs, we search through a decoding graph similar to \cite{sainath17} but accept both phonemes and wordpieces. An example is shown in Figure \ref{fig:decoding_fst}. The decoding FST has wordpiece loops around state 0 (we show only a few for simplicity), but also has a pronunciation section (states 1 through 14).
The pronunciation section is a prefix tree with phonemes as inputs, and outputs are wordpieces of the corresponding word produced by the WPM in Section \ref{sec:wpp}.
Specifically, for each word in the biasing list, we look up pronunciations from the lexicon and split the word into its constituent wordpieces. Input phoneme labels are accepted and transduced into wordpieces. Input wordpiece labels are accepted by the wordpiece loops. The final output symbols, which are always wordpieces, are concatenated into words.

Based on \cite{sainath17}, we add two improvements to the decoding strategy.
First, during decoding we consume as many input epsilon arcs as possible thus guaranteeing that all wordpieces in word are produced when all corresponding phonemes are seen in the input.
Second, we merge paths that have the same \textit{output} symbols. Given the nature of our training and decoding, a given word can be output either directly in wordpieces, or transduced from phonemes to wordpieces. Since the input symbols are different, each hypothesis has a different probability. We keep track of equivalent hypotheses and recombine them by adding their probabilities, assigning the total probability to the most likely hypothesis, and dropping the others from the beam.

\begin{table*}[h]
\centering
\begin{tabular}{ |l|l|c|c|c| }
    \hline
    Model & Biasing unit & \emph{Directions} w/o Bias & \emph{Directions} w/ Bias & General Traffic (English) \\ \hline
    \hline
    Grapheme & Graphemes & 36.0 & 32.8 & 5.9\\ \hline
    WPM & Wordpieces & 38.3 & 30.0 & 5.6 \\ \hline
    Wordpiece-Phoneme & Phonemes & 35.9 & 27.5 & 5.7  \\ \hline
    \quad+ wordpiece biasing & Wordpieces and phonemes & 35.9 & 26.9 & 5.7 \\ \hline
\end{tabular}
\caption{Comparison of WERs (\%) of different models for a foreign place name recognition task and Google general English traffic.}
\label{tab:wer}
\vspace{-1em}
\end{table*}

\section{Experiments}
\label{sec:exp}

\subsection{Data Sets}
\label{sec:data}
Our training set contains 35 million English utterances with a total of around 27,500 hours. These utterances are sampled from Google's general English traffic, and are anonymized and hand-transcribed for training. To increase training diversity, clean utterances are artificially corrupted by using a room simulator, varying degrees of noise, and reverberation such that the overall signal-to-noise ratio (SNR) is between 0dB and 30dB, and an average SNR is 12dB \cite{kim17}. The noise sources are from YouTube and daily life noisy environmental recordings. 

Utterances with cross-lingual words are hardly present in our data set, and thus we use a TTS engine, parallel-wavenet \cite{oord17}, to synthesize utterances for evaluation. We choose French as the foreign language, and the utterances consist of navigation queries (e.g. ``directions to Créteil"). There are in total 1K utterances and we refer to this set as the \emph{Directions} test set. For each utterance, the bias set contains 1K words including the ground-truth place name and unrelated French place names. Since all biasing words are in a foreign language, they have never been seen in training. In decoding, all biasing words are used to construct a contextual FST with each arc having the same weight. In later evaluation, this weight is tuned independently for different models.

On the other hand, to evaluate how the wordpiece-phoneme model performs on the regular English recognition task, we sampled a total of 30.5K English utterances from general Google traffic as the no-biasing test set.

\subsection{Model \replaced{Training}{Architecture}}
Similarly to \cite{he18}, an input utterance is divided to 25-ms frames, windowed and shifted at a rate of 10 ms. A 80-dimensional log-Mel feature is extracted at each frame, and the current frame and two frames to the left are concatenated to produce a 240-dimensional log-Mel feature. These features are then downsampled at a rate of 30 ms. We use RNN-T as the sequence-to-sequence model. Similar to \cite{he18}, the encoder of the RNN-T consists of 8 Long Short-Term Memory (LSTM) \cite{hochreiter1997} layers and the prediction network contains 2 layers. Each LSTM layer contains 2,048 hidden units followed by a 640-dimensional projection layer. A time-reduction layer is added after the second layer to improve the inference speed without accuracy loss. Outputs of the encoder and prediction network are fed to a joint-network which has 640 hidden units, which is then followed by a softmax layer containing 4,096 output units. Specifically, the output units contain 41 context-independent phonemes and the rest are wordpieces.

As described in Section \ref{sec:wpp}, we use a lexicon containing about 430,000 words with their frequencies to determine when to use phoneme sequences. \added{The lexicon contains words and their frequencies from training data, and is trimmed by removing homophones (e.g. ``flower" and ``flour"), homographs (e.g. ``live" as a verb or adjective), and pronunciation variants (e.g. ``either"). It thus only contains entries that are unambiguous when going from spelling to pronunciation or the other way around. We do not generate phonemes for out-of-lexicon words using a trained grapheme-to-phoneme model. The intuition is that when pronunciation is not known, it is simpler and cleaner to let the E2E model infer the pronunciation rather than bring in another independently trained model. In addition, we use the written form of a transcript and do not use any verbalizer. Thus, words like ``\$9.95" were never presented as phonemes. In model training,} a <$\text{eow}$> symbol is inserted between words to identify spacing. \deleted{Since a phoneme sequence tends to be longer than a wordpiece sequence, we use a maximum length of 120 output tokens in the target sequence. }The model contains around 120M parameters in total. All RNN-T models are trained in Tensorflow \cite{abadi2016tensorflow} on 8 $\times$ 8 Tensor Processing Units (TPU) slices with a global batch size of 4,096.

\subsection{WERs and Comparisons}

We compare the biasing results of the wordpiece-phoneme model to a grapheme-only model and a wordpiece-only model. The latter two models have the same structure as the wordpiece-phoneme model. The difference is that the grapheme model has 76 graphemes as outputs and the wordpiece model has 4,096 wordpieces. This leads to around 117M and 120M parameters for the grapheme model and wordpiece model, respectively. Note that the two model's output symbols are in English and they are trained using all-English data described in Section \ref{sec:data}. \TB{Rohit says to talk about how diacritics are handled.} \KH{Done} For these two models, biasing is done at the grapheme level or wordpiece level alone \added{using the English transliterated versions of French biasing words}. WERs of the \emph{Directions} set are shown in Table \ref{tab:wer}.

First, we see that all three models perform poorly without biasing. This is because the place names are in French and they have never been seen in training, i.e. an word OOV rate of nearly 100\%\footnote{Note that we transliterate the ground truth transcripts of the \emph{Directions} set to English for WER computation for the grapheme and wordpiece models since the they cannot produce words with French diacritics.}. Secondly, we see in Table \ref{tab:wer} that all models performs substantially better with biasing. The WER reductions range from 9\%-23\% relatively for different models when compared to the no-bias case. Comparing different biasing strategies, we find that the wordpiece-phoneme model performs the best: 16\% relatively better than the grapheme model, and 8.3\% better than the wordpiece model. We attribute the superior performance of the wordpiece-phoneme model to the robustness of phonemes to OOV words, as observed in \cite{sainath17}.

Since the wordpiece-phoneme model contains both wordpieces and phonemes as modeling units, we can further perform wordpiece biasing in addition to phoneme-based biasing by building a wordpiece FST in parallel to the phoneme FST. This further reduces the WER by 2\%, as shown in the bottom row in Table \ref{tab:wer}. This shows that wordpiece and phoneme biasing are complementary to each other. We note that the same weights are used for both phoneme and wordpiece biasing, and empirically we did not find significant improvements by using different weights. On the other hand, for wordpiece model biasing, our results are consistent with the observation in \cite{zhao19} that the wordpieces perform better than graphemes because of its sparsity in matching longer units.

To further understand how biasing helps recognizing French place names, we present some wins of wordpiece-phoneme model in Table \ref{tab:win}. We can see that biasing helps produce the correct French words, and in contrast, phonetically similar but wrong English words are produced when without biasing. On the other hand, we present some typical recognition errors in Table \ref{tab:error}. We see that errors are mainly due to phonetically similar words in French. We will analyze how the biasing performance changes as the number of biasing words changes in Section \ref{sec:distractor}.

To ensure there is no regression in no-biasing scenarios, we compare three models in decoding regular English utterances from general Google traffic. In decoding, we turn the biasing mechanism off by using an empty list of biasing phrases. As shown in the last column of Table \ref{tab:wer}, the wordpiece model performs better than the grapheme model as in \cite{he18}. The wordpiece-phoneme model performs a little better than the grapheme model, and we attribute that to \replaced{the higher frequency of wordpieces during training}{dominating wordpieces in the model}. Compared to the wordpiece model, the wordpiece-phoneme model has a slight degradation (0.1\% absolute WER). This is due to the introduction of phonemes in modeling. One potential approach to improve regression is to incorporate an English external language model for phonemes in rescoring, similarly to the wordpiece-based rescoring in \cite{kannan18}.  
However, we note that the regression is significantly smaller than the all-phoneme model in \cite{sainath17} \TB{what was the degradation?}. \KH{I tend not to mention an exact number since setup is pretty different.}

\begin{table}
\centering
\begin{tabular}{ |c|c| }
    \hline
     w/ Bias (correct) & w/o Bias \\ \hline
      directions to Champs-Élysées & directions to shaw city  \\ \hline
      directions to Bouches-du-Rhône & directions to bushnell \\ \hline
      directions to Ardèche & directions to adesh \\ \hline
\end{tabular}
\caption{Comparison of correctly recognized French words (w/ biasing) and errors (w/o biasing).}
\label{tab:win}
\vspace{-1em}
\end{table}

\begin{table}
\centering
\begin{tabular}{ |c|c| }
    \hline
    Estimated & Reference \\ \hline
      directions to Citroën & directions to Saint-Honoré\\ \hline
      directions to Marne-la-Vallée & directions to Marcq-en-Barœul \\ \hline
      directions to Métropole & directions to Megève \\ \hline
\end{tabular}
\caption{Examples of wrongly recognized French words.}
\label{tab:error}
\vspace{-1em}
\end{table}

\begin{figure}[h!]
  \centering
  \includegraphics[scale=0.27]{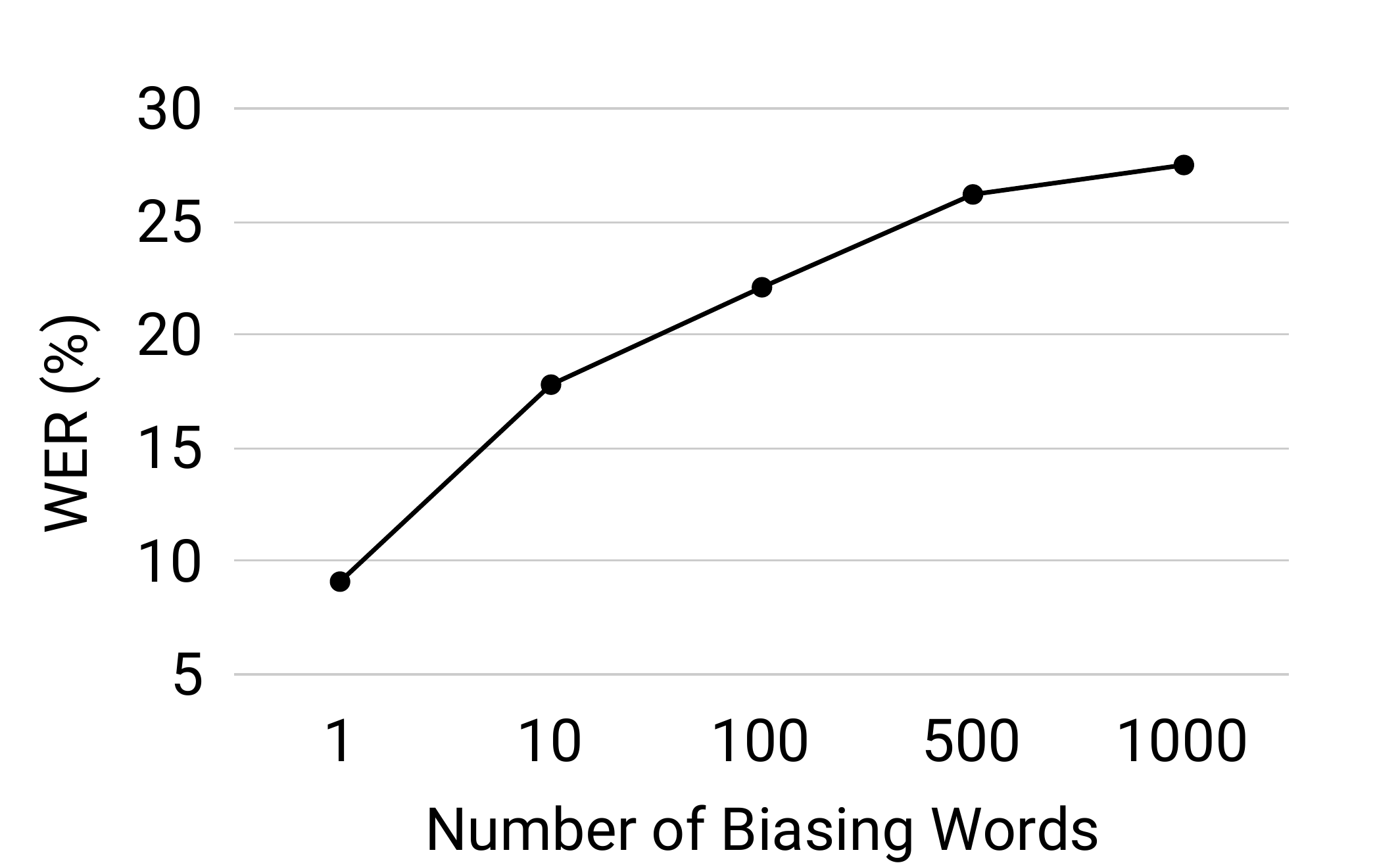}
   \caption{{WER (\%) as a function of the number of biasing words.}}
   \label{fig:distractors}
   \vspace{-1em}
\end{figure}

\subsection{Effect of Number of Biasing Words}
\label{sec:distractor}

Given examples in Table \ref{tab:error}, we are curious how competing biasing words affect recognition. We thus randomly choose a fixed number of biasing words (including the ground truth one) and vary the number to see how WER changes. Figure \ref{fig:distractors} shows that WER for the \emph{Directions} set is 9.1\% when only the ground truth word is present (i.e. number of biasing words is 1), and the rate increases quickly when the total number of biasing words increases. We attribute the quick degradation to the significant matching confusion in the phoneme prefixes of the words as the number of biasing words increases (as confirmed by phonetically similar French place names in Table \ref{tab:error}). One interesting direction would be to increase the length of the phonemic units to create a sparser match.

\section{Conclusion}
\label{sec:conclude}

In this work we proposed a wordpiece-phoneme RNN-T model and phoneme-level contextual biasing to recognize foreign words. Biasing at the phoneme level enables us to avoid the OOV problem in the wordpiece model. Evaluating on a test set containing navigation queries to French place names, we show the proposed approach performs significantly better than a state-of-the-art grapheme and wordpiece model, by 16\% and 8\%, respectively in terms of relative WER reductions. Wordpiece biasing is complimentary to phoneme biasing and adds a further 2\% reduction.
\deleted{In this work, we do not address biasing homophones, and one solution would be to better leverage prefixes and wordpieces in biasing. }Lastly, since wordpieces perform better than graphemes \cite{he18} in E2E modeling, it would be interesting to explore longer phonemic units such as phoneme pieces for biasing.

\bibliographystyle{IEEEtran}

\bibliography{mybib}

\begin{thebibliography}{10}
\providecommand{\url}[1]{#1}
\csname url@samestyle\endcsname
\providecommand{\newblock}{\relax}
\providecommand{\bibinfo}[2]{#2}
\providecommand{\BIBentrySTDinterwordspacing}{\spaceskip=0pt\relax}
\providecommand{\BIBentryALTinterwordstretchfactor}{4}
\providecommand{\BIBentryALTinterwordspacing}{\spaceskip=\fontdimen2\font plus
\BIBentryALTinterwordstretchfactor\fontdimen3\font minus
  \fontdimen4\font\relax}
\providecommand{\BIBforeignlanguage}[2]{{%
\expandafter\ifx\csname l@#1\endcsname\relax
\typeout{** WARNING: IEEEtran.bst: No hyphenation pattern has been}%
\typeout{** loaded for the language `#1'. Using the pattern for}%
\typeout{** the default language instead.}%
\else
\language=\csname l@#1\endcsname
\fi
#2}}
\providecommand{\BIBdecl}{\relax}
\BIBdecl

\bibitem{soltau16}
H.~Soltau, H.~Liao, and H.~Sak, ``Neural speech recognizer: Acoustic-to-word
  {LSTM} model for large vocabulary speech recognition,'' \emph{arXiv preprint
  arXiv:1610.09975}, 2016.

\bibitem{graves12}
A.~Graves, ``Sequence transduction with recurrent neural networks,''
  \emph{arXiv preprint arXiv:1211.3711}, 2012.

\bibitem{rao17}
K.~Rao, H.~Sak, and R.~Prabhavalkar, ``Exploring architectures, data and units
  for streaming end-to-end speech recognition with rnn-transducer,'' in
  \emph{2017 IEEE Automatic Speech Recognition and Understanding Workshop
  (ASRU)}.\hskip 1em plus 0.5em minus 0.4em\relax IEEE, 2017, pp. 193--199.

\bibitem{chan16}
W.~Chan, N.~Jaitly, Q.~Le, and O.~Vinyals, ``Listen, attend and spell: A neural
  network for large vocabulary conversational speech recognition,'' in
  \emph{Acoustics, Speech and Signal Processing (ICASSP), 2016 IEEE
  International Conference on}.\hskip 1em plus 0.5em minus 0.4em\relax IEEE,
  2016, pp. 4960--4964.

\bibitem{chiu18}
C.-C. Chiu, T.~N. Sainath, Y.~Wu, R.~Prabhavalkar, P.~Nguyen, Z.~Chen,
  A.~Kannan, R.~J. Weiss, K.~Rao, E.~Gonina \emph{et~al.}, ``State-of-the-art
  speech recognition with sequence-to-sequence models,'' in \emph{2018 IEEE
  International Conference on Acoustics, Speech and Signal Processing
  (ICASSP)}.\hskip 1em plus 0.5em minus 0.4em\relax IEEE, 2018, pp. 4774--4778.

\bibitem{he18}
\BIBentryALTinterwordspacing
Y.~He, T.~N. Sainath, R.~Prabhavalkar, I.~McGraw, R.~Alvarez, D.~Zhao,
  D.~Rybach, A.~Kannan, Y.~Wu, R.~Pang \emph{et~al.}, ``Streaming end-to-end
  speech recognition for mobile devices,'' \emph{ICASSP}, 2019. [Online].
  Available: \url{\url{https://arxiv.org/pdf/1811.06621.pdf}}
\BIBentrySTDinterwordspacing

\bibitem{aleksic15}
P.~Aleksic, M.~Ghodsi, A.~Michaely, C.~Allauzen, K.~Hall, B.~Roark, D.~Rybach,
  and P.~Moreno, ``Bringing contextual information to {Google} speech
  recognition,'' in \emph{Sixteenth Annual Conference of the International
  Speech Communication Association}, 2015.

\bibitem{hall15}
K.~Hall, E.~Cho, C.~Allauzen, F.~Beaufays, N.~Coccaro, K.~Nakajima, M.~Riley,
  B.~Roark, D.~Rybach, and L.~Zhang, ``Composition-based on-the-fly rescoring
  for salient n-gram biasing,'' 2015.

\bibitem{williams18}
I.~Williams, A.~Kannan, P.~Aleksic, D.~Rybach, and T.~N. Sainath, ``Contextual
  speech recognition in end-to-end neural network systems using beam search,''
  \emph{Proc. Interspeech 2018}, pp. 2227--2231, 2018.

\bibitem{kannan18}
A.~Kannan, Y.~Wu, P.~Nguyen, T.~N. Sainath, Z.~Chen, and R.~Prabhavalkar, ``An
  analysis of incorporating an external language model into a
  sequence-to-sequence model,'' in \emph{2018 IEEE International Conference on
  Acoustics, Speech and Signal Processing (ICASSP)}.\hskip 1em plus 0.5em minus
  0.4em\relax IEEE, 2018, pp. 1--5828.

\bibitem{pundak18}
G.~Pundak, T.~N. Sainath, R.~Prabhavalkar, A.~Kannan, and D.~Zhao, ``Deep
  context: end-to-end contextual speech recognition,'' \emph{arXiv preprint
  arXiv:1808.02480}, 2018.

\bibitem{schuster2012japanese}
M.~Schuster and K.~Nakajima, ``Japanese and korean voice search,'' in
  \emph{2012 IEEE International Conference on Acoustics, Speech and Signal
  Processing (ICASSP)}.\hskip 1em plus 0.5em minus 0.4em\relax IEEE, 2012, pp.
  5149--5152.

\bibitem{zhao19}
D.~Zhao, T.~N. Sainath, D.~Rybach, D.~Bhatia, B.~Li, and R.~Pang,
  ``Shallow-fusion end-to-end contextual biasing,'' \emph{To appear in
  Interspeech 2019}, 2019.

\bibitem{bruguier19}
A.~Bruguier, R.~Prabhavalkar, G.~Pundak, and T.~N. Sainath, ``Phoebe:
  {P}ronunciation-aware contextualization for end-to-end speech recognition,''
  in \emph{to appear in Proc. ICASSP, 2019}.\hskip 1em plus 0.5em minus
  0.4em\relax IEEE.

\bibitem{patel18}
A.~Patel, D.~Li, E.~Cho, and P.~Aleksic, ``Cross-lingual phoneme mapping for
  language robust contextual speech recognition,'' in \emph{2018 IEEE
  International Conference on Acoustics, Speech and Signal Processing
  (ICASSP)}.\hskip 1em plus 0.5em minus 0.4em\relax IEEE, 2018, pp. 5924--5928.

\bibitem{sainath17}
T.~N. Sainath, R.~Prabhavalkar, S.~Kumar, S.~Lee, A.~Kannan, D.~Rybach,
  V.~Schogol, P.~Nguyen, B.~Li, Y.~Wu \emph{et~al.}, ``No need for a lexicon?
  {E}valuating the value of the pronunciation lexica in end-to-end models,''
  \emph{ICASSP}, 2017.

\bibitem{irie19}
K.~Irie, R.~Prabhavalkar, A.~Kannan, A.~Bruguier, D.~Rybach, and P.~Nguyen,
  ``Model unit exploration for sequence-to-sequence speech recognition,''
  \emph{arXiv preprint arXiv:1902.01955}, 2019.

\bibitem{bruguier17}
A.~Bruguier, D.~Gnanapragasam, L.~Johnson, K.~Rao, and F.~Beaufays,
  ``Pronunciation learning with {RNN}-transducers,'' \emph{Proc. Interspeech
  2017}, pp. 2556--2560, 2017.

\bibitem{kim17}
C.~Kim, A.~Misra, K.~Chin, T.~Hughes, A.~Narayanan, T.~Sainath, and
  M.~Bacchiani, ``Generation of large-scale simulated utterances in virtual
  rooms to train deep-neural networks for far-field speech recognition in
  google home,'' 2017.

\bibitem{oord17}
A.~v.~d. Oord, Y.~Li, I.~Babuschkin, K.~Simonyan, O.~Vinyals, K.~Kavukcuoglu,
  G.~v.~d. Driessche, E.~Lockhart, L.~C. Cobo, F.~Stimberg \emph{et~al.},
  ``Parallel wavenet: Fast high-fidelity speech synthesis,'' \emph{arXiv
  preprint arXiv:1711.10433}, 2017.

\bibitem{hochreiter1997}
S.~Hochreiter and J.~Schmidhuber, ``Long short-term memory,'' \emph{Neural
  computation}, vol.~9, no.~8, pp. 1735--1780, 1997.

\bibitem{abadi2016tensorflow}
M.~Abadi, P.~Barham, J.~Chen, Z.~Chen, A.~Davis, J.~Dean, M.~Devin,
  S.~Ghemawat, G.~Irving, M.~Isard \emph{et~al.}, ``Tensorflow: A system for
  large-scale machine learning,'' in \emph{12th {USENIX} Symposium on Operating
  Systems Design and Implementation ({OSDI} 16)}, 2016, pp. 265--283.

\end{thebibliography}

\end{document}